\documentclass[lettersize,journal]{IEEEtran}
\usepackage{amsmath,amsfonts}
\usepackage{algpseudocode}
\usepackage{algorithm}
\usepackage{array}
\usepackage[caption=false,font=normalsize,labelfont=sf,textfont=sf]{subfig}
\usepackage{textcomp}
\usepackage{stfloats}
\usepackage{url}
\usepackage{verbatim}
\usepackage{xcolor}
\usepackage{CJKutf8}
\usepackage{graphicx}
\usepackage{cite}
\usepackage{booktabs}
\usepackage{multirow}
\usepackage{tabularx}
\begin{document}
\bstctlcite{IEEEtranBSTCTL}

\title{Cross-Protocol Personality Assessment with LLMs: Aligning Behavioral Evidence with Human Ratings}

\author{
Wei Shan,
Tianyi~Zhang$^{*}$,~\IEEEmembership{Senior Member,~IEEE,}
Yuan~Zong, ~\IEEEmembership{Member,~IEEE,}
and~Wenming~Zheng, ~\IEEEmembership{Senior Member,~IEEE,}
\thanks{Wei Shan, Tianyi Zhang, Yuan Zong, and Wenming Zheng are with the Key Laboratory of Child Development and Learning Science (Ministry of Education), School of Biological Sciences and Medical Engineering, Southeast University, Nanjing, China.}
\thanks{Corresponding author: Tianyi Zhang: {t.zhang@seu.edu.cn}}
}
\markboth{IEEE Transactions on Multimedia}{IEEE TMM}

\maketitle

\begin{abstract}
Personality assessment from asynchronous video interviews (AVIs) plays a crucial role in recruitment by enabling standardized and behaviorally grounded evaluation for job applicants. However, most of the previous works are confined to a single interview protocol. Independently collected protocols differ in question sets and question-trait mapping. Thus, the human-rated evidence from one protocol cannot be used directly under another. To address this limitation, we propose the Cross-protocol Evidence-grounded Personality Assessment (CEPA) framework, which aligns behavioral representations across interview protocols. CEPA addresses protocol heterogeneity by projecting source behavioral records into the target protocol’s construct-oriented representation space. It retrieves behaviorally relevant and personality-complementary human-rated exemplars to ground an ensemble of LLM raters, which enables behavior-to-personality associations without target-specific supervision.
In cross-protocol experiments, CEPA reduces average mean squared error (MSE) by 43.8\% (Forward) and 17.0\% (Reverse) relative to the strongest baseline. Our empirical experiments show that human-rated evidence from one protocol can support personality assessment under another without target-specific supervision or retraining.
\end{abstract}

\section{Introduction}
\label{sec:intro}

Asynchronous video interviews (AVIs) are now a widely used tool in high-volume personnel screening. In AVIs, candidates record answers to predefined and structural questions for interviewer screening~\cite{brenner2016,suen2019}. The structured format scales to large candidate pools at low cost. Many companies use AVIs to assess personality traits in job recruitment as personality traits are considered as important indicators of how well someone might fit the job and workplace culture~\cite{hickman2022automated,liff2024psychometric}. Thus, plenty of multimodal algorithms have been developed to assess personality traits from AVIs ~\cite{koutsoumpis2024,mwcf2023,avi6}. 


The uniqueness of AVI-based personality assessment is that behavior cues related to personality traits are activated by structured and protocol-specific AVI questions. According to Trait Activation Theory~\cite{tett2003}, situational cues (i.e., questions in AVIs) trigger the expression of relevant latent traits through observable behaviors. Unlike personality assessment from daily conversations or vlog~\cite{wei2018,li2020,wang2018} (i.e., first impression personality), the models for AVI-based personality assessment need to capture the question-trait mapping and the correspondence between each question and the traits it activated. 


Although previous works have developed plenty of methods to capture the question-trait mappings, they are typically restricted to a single AVI protocol. Independently collected protocols are heterogeneous in their questions and question-trait mappings even when they target the same traits. A question may target one trait or several, and multiple questions can target the same trait. Thus, the same observed behavior can carry different trait interpretations under different protocols. For example, a candidate’s description of organizing a team project might be interpreted as evidence of Extraversion under one protocol and Conscientiousness under another. The cross-protocol problem can lead to the cross-protocol problem for AVI-based personality assessment: behavioral evidence gathered under one protocol cannot be used directly under another.

To address the cross-protocol problem, two kinds of methods are developed. Some of the methods use zero-shot LLM to assess personality from behavioral language without target-specific supervision. Its judgments agree with self-report ratings and are valid in AVI-based assessment~\cite{peters2024,wright2026,ganesan2023,ji2023,zhang2024}. However, this kind of method ignores the fact that the human judgments already collected under other protocols. Other methods use supervised fine-tuning, including parameter-efficient methods such as LoRA, adapts a model to a target protocol~\cite{shen2025,zhang2026}. This kind of method needs target-specific supervision and optimization for every new protocol. Neither approach therefore exploits evidence already collected under other protocols.

To address this limitation, we develop \textbf{CEPA}, Cross-protocol Evidence-grounded Personality Assessment, a retrieval-augmented framework for multimodal personality assessment across protocols. CEPA can reuse human-rated behavioral evidence from existing protocols to assess personality under new protocols without requiring labeled target data or model retraining. CEPA has two modules: the Representation Alignment module and the personality-space Max-Min Evidence Selection module. \textbf{Representation Alignment} maps multimodal behavioral records from different protocols into a shared representation that preserves each record's personality profile. \textbf{Personality-space Max-Min Evidence Selection (PMES)} then constructs a compact evidence pool that balances behavioral relevance with coverage of the personality space. Finally, multiple LLM raters reason over the retrieved records and their judgments are aggregated into a continuous personality score. Together, these components enable human-rated records from one protocol to provide empirical evidence for personality assessment under another protocol. In general, our work make below contributions to the multimedia community:
\begin{enumerate}
\item We propose \textbf{CEPA}, a protocol-agnostic framework that reuses human-rated records from one assessment protocol to score personality under another. It relies on previously collected human judgments at \textbf{assessment} time and requires no target-specific supervision or optimization. This avoids the evidence loss of zero-shot assessment and the per-protocol supervision that fine-tuning requires.

\item In bidirectional transfer between two independently collected AVI datasets, CEPA reduces average MSE by $43.8\%$ (Forward) and $17.0\%$ (Reverse) over the strongest zero-shot baseline. Ablations attribute most of the gain to retrieved human-rated evidence and Representation Alignment.

\item We show that the empirical relationship between observed behavior and human personality judgment transfers across protocols. By contrast, the source personality label does not transfer. Using retrieved records as evidence consistently outperforms using their attached personality profiles directly. A multi-rater reliability analysis separates prediction error from the stability of the scoring process.
\end{enumerate}

\section{Related Work}
\label{sec:related}


\subsection{Personality Assessment from Asynchronous Video Interviews}

Early work predicted the Sixteen Personality Factor (16PF) dimensions from facial features~\cite{gavrilescu2017}. Deep bimodal regression predicted apparent personality from short video sequences~\cite{wei2018}. Later models combined channels through trait-specific fusion and joint classification-regression objectives~\cite{li2018,li2020,zhao2023}. Multimodal fusion now dominates because personality evidence is spread across verbal and non-verbal behavior~\cite{song2023,bao2024,duan2024,mwcf2023}. These fusion models are tied to the protocol they were trained on. Their behavior-to-trait mappings do not transfer to new question sets or trait definitions. Large language models instead score personality directly from behavioral language.

A first line of LLM-based work applies pre-trained models in a zero-shot setting. Peters and Matz~\cite{peters2024} showed that GPT-3.5 and GPT-4 infer Big Five traits from Facebook updates. Ganesan et al.~\cite{ganesan2023} found that zero-shot GPT-3 matched a supervised state-of-the-art model on broad Big Five classification once trait knowledge was added to the prompt, but fell to a most-frequent-class baseline on fine-grained classification. Zhang et al.~\cite{zhang2024} provide a systematic validity and fairness evaluation of LLM-based AVI assessment. These methods score a candidate from observed behavior alone. They never reuse human-rated records collected under other protocols. The cross-protocol problem is left unaddressed.

A second line adapts LLMs with target-specific supervision. Representative methods fine-tune the LLM~\cite{shen2025} or train a predictor over its embeddings~\cite{zhang2026-fau-fusion}. Zhang et al.~\cite{zhang2026} proposed PersonalityLLM, a coarse-to-fine LoRA framework that regresses a continuous score within an empirical trait range, improving prediction by $16.6\%$/$37.9\%$ on two benchmarks. MoA-Personality~\cite{zhang-etal-2026-moe-personality} pools multiple LLM annotators with an aggregator to curb positivity bias, outperforming medium-sized open-source LLMs and approaching closed-source GPT-3.5 and GPT-4.

These methods learn a protocol-specific behavior-to-trait mapping from annotated data. Neither line reuses evidence from other protocols. The zero-shot line discards prior human judgments. The fine-tuning line needs target-specific supervision for each new protocol. Retrieval-augmented generation avoids per-protocol training by conditioning on retrieved evidence.

\subsection{Retrieval-Augmented Generation and Evidence Transfer}

Retrieval-augmented generation (RAG) combines a parametric language model with a non-parametric retrieval index. Karpukhin et al.~\cite{karpukhin2020} showed that neural dual-encoders can replace sparse retrievers, and Lewis et al.~\cite{lewis2020} built a general framework around a dense retriever and a parametric generator. Retrieval has since been integrated into language-model pre-training~\cite{guu2020,borgeaud2022}, with the RAG design space now covered by dedicated surveys~\cite{gao2023}. Effective evidence selection requires relevant items and a diverse retrieved set~\cite{carbonell1998,yang2023,khan2026,levy2023}.

Personality assessment imposes a different evidence requirement. RAG in open-domain QA retrieves factual passages that contain an answer. Personality assessment instead needs records that tie observed behavior to a human judgment of the trait. Conventional RAG assumes a general or task-specific knowledge base. No such base links multimodal behavior to traits across different question-trait mappings. Independently collected datasets fill this role. Because they already pair behavioral observations with human personality judgments. These knowledgeable-other judgments have well-documented validity~\cite{kolar1996}. One dataset can serve as an evidence base for another.

This setting connects to cross-dataset transfer and domain adaptation. Classic methods learn domain-invariant features through gradient reversal~\cite{pmlr-v37-ganin15} or RKHS mean matching~\cite{pmlr-v37-long15}. Adaptation has also been applied to personality analysis across short-video platforms~\cite{an2024personalityanalysisonlineshort} and to RAG retrieval for open-domain QA~\cite{siriwardhana2023}. These methods share one assumption: each transfers a model or a feature representation. Retrieval-based methods adjust a retrieval policy. They usually require target-domain data or parameter updates. They do not address whether the behavior-to-trait mappings embedded in human-rated records can themselves be transferred at assessment time. CEPA targets this gap directly. It retrieves human-rated records from one protocol as an evidence base and aggregates the judgments of multiple LLM raters. This enables behavior-to-trait mappings to transfer across protocols without target-specific supervision or parameter updates.

\section{Method}
\label{sec:method}

\subsection{Problem Setting and Framework Overview}

We consider cross-protocol transfer of multimodal behavioral evidence for personality assessment. The independently collected interview datasets may differ in their question-trait mappings. Let $\mathcal{D}_s$ and $\mathcal{D}_t$ denote the source and target datasets, respectively. The source provides human-rated behavioral exemplars for knowledge-base (KB) construction. The target is held out from personality-rating supervision and from retrieval-configuration selection. Candidate-pool size, evidence budget, and retrieval encoder are not tuned on the target. At assessment, only the target subject's behavioral observations are available as a query.

Let $\mathcal{F}_t$ denote the construct space induced by the target assessment protocol and used for evidence construction, selection, and scoring. $\mathcal{F}_t$ is defined by the target protocol, so its granularity may differ from the source dataset's native annotation granularity. During KB construction, each source profile is mapped into $\mathcal{F}_t$. Every admitted evidence card then carries a complete target-space profile.

\begin{figure*}[t]
\centering
\includegraphics[width=\linewidth]{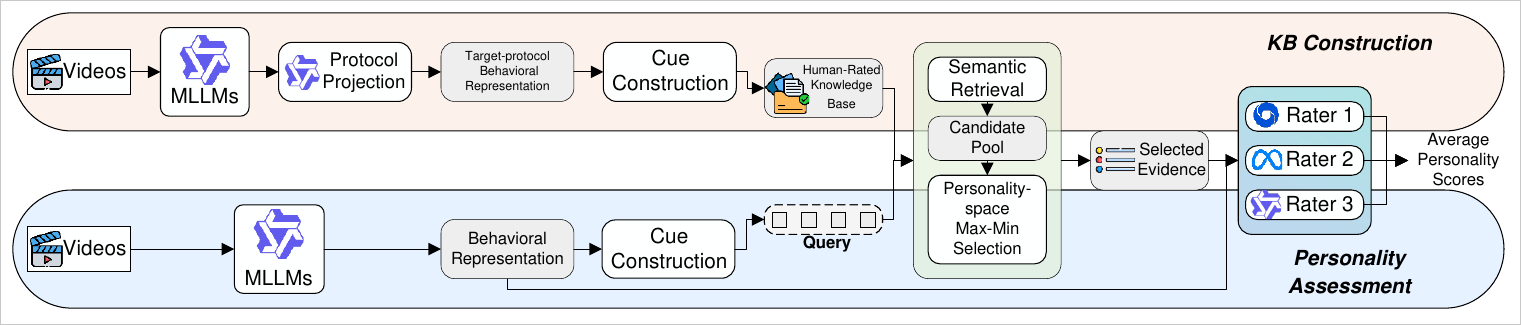}
\caption{Overview of the proposed \textbf{CEPA} framework. Multimodal interview videos are first converted into textual behavioral representations. Source evidence is aligned to the target protocol and encoded into a shared construct-oriented retrieval space. For a target subject, semantic retrieval forms a candidate pool, PMES selects complementary human-rated exemplars, and independent LLM raters produce continuous personality scores.}
\label{fig:pipeline}
\end{figure*}

We aim to transfer human-rated behavioral evidence from $\mathcal{D}_s$ to $\mathcal{D}_t$ and predict continuous personality scores without target-dataset personality supervision. Fig.~\ref{fig:pipeline} shows the proposed framework, which we call \textbf{CEPA} (Cross-protocol Evidence-grounded Personality Assessment). It has three stages: \textbf{Representation Alignment}, \textbf{Personality-space Max-Min Evidence Selection (PMES)}, and \textbf{Evidence-Grounded Continuous Scoring}. Representation Alignment conditions source behavior on the target protocol and converts it into a construct-oriented retrieval representation. PMES retrieves a relevant candidate region by semantic similarity and then selects complementary human-rated exemplars. Evidence-Grounded Continuous Scoring uses independent LLM raters to produce continuous personality scores conditioned on the selected evidence.

\subsection{Multimodal Behavioral Preprocessing}
\label{sec:preprocessing}

Each interview video is first converted into a textual multimodal behavioral representation before evidence construction or assessment. The preprocessing stage obtains the interview transcription together with textual descriptions of audio and visual behaviors. These descriptions form a modality-serialized representation of the observable interview behavior and serve as input to the Protocol Projection and Cue Construction modules.

We denote the resulting behavioral representations of a source subject $i$ and of the target subject by $x_i$ and $x_t$, respectively. The preprocessing procedure is applied independently to the source and target data and does not use personality ratings from the target dataset.

\subsection{Representation Alignment}
\label{sec:representation_alignment}

Representation Alignment transforms source behavioral evidence into a target-protocol-compatible retrieval representation and preserves the source human-rated profile. It has two sequential operations. \emph{Protocol Projection} aligns source behavior with the target protocol. \emph{Cue Construction} converts the projected behavior into a construct-oriented semantic representation.

\subsubsection{Protocol Projection}

Let $P_s$ and $P_t$ denote the source and target assessment protocols. For source behavioral evidence $x_i$ with human-rated profile $y_i^{s}$, Protocol Projection constructs the target-protocol-compatible representation $\tilde{x}_{i\rightarrow t}$. It also maps the source profile into the target construct space as $y_{i\rightarrow t} = (s_f^i)_{f\in\mathcal{F}_t}$, where $s_f^i$ is the human-rated score of subject $i$ on construct $f$. The projection rewrites only the behavioral representation. The human-rated profile comes from source annotations through a predefined construct mapping and is not inferred from the projected behavior.

The target construct space is $\mathcal{F}_t=\{f_1,\ldots,f_{|\mathcal{F}_t|}\}$, and $\mathcal{C}_f$ denotes the official definition and assessment description of construct $f$. The source-to-target profile mapping provides a human-rated score $s_f^i$ for each $f\in\mathcal{F}_t$ required by the transfer setting. Source records without a complete mapped profile are excluded from the KB rather than having missing construct scores imputed. All evidence cards admitted to PMES carry a complete human-rated profile in the target construct space.

Projection is applied during KB construction. Projected records are admitted only if they pass structural-completeness and score-leakage checks. All $|\mathcal{F}_t|$ construct slots must be populated, and the projected representation must not contain the numeric score value. Records that fail either check are discarded rather than repaired.

\subsubsection{Cue Construction and Semantic Representation}

Each projected behavioral representation is converted into a
construct-oriented textual cue:

\begin{equation}
c_i^f =
G_{\mathrm{cue}}(\tilde{x}_{i\rightarrow t},\mathcal{C}_f),
\label{eq:cue}
\end{equation}
where $G_{\mathrm{cue}}(\cdot)$ denotes the cue-generation template and
$c_i^f$ is the resulting textual cue of subject $i$ on construct $f$.

Each cue organizes observable evidence into five semantic slots: \textsc{high-pole evidence}, \textsc{low-pole evidence}, \textsc{transcript evidence}, \textsc{audio evidence}, and \textsc{video evidence}. The cue does not emit numerical personality scores or diagnostic trait labels. Interview questions are not reproduced, and slot labels are removed before embedding. Prompt specifications and validation constraints are provided in the Supplementary Material.

For KB construction, source cues are generated from projected source representations. Protocol Projection maps source multimodal observations into target-protocol answer fields, so the \textsc{audio evidence} and \textsc{video evidence} slots are set to \emph{none}. Target cues are generated directly from target transcript, audio, and video descriptions. Both sides use the same construct definitions, score-free output format, semantic encoder, and per-construct aggregation.

Let $E_{\mathrm{cue}}(\cdot)$ denote the semantic embedding operation. The subject-level representations are

\begin{equation}
\begin{aligned}
z_i &=
\frac{1}{|\mathcal{F}_t|}
\sum_{f\in\mathcal{F}_t}
E_{\mathrm{cue}}\left(
G_{\mathrm{cue}}\left(
\tilde{x}_{i\rightarrow t},\mathcal{C}_f
\right)
\right),\\
z_t &=
\frac{1}{|\mathcal{F}_t|}
\sum_{f\in\mathcal{F}_t}
E_{\mathrm{cue}}\left(
G_{\mathrm{cue}}\left(
x_t,\mathcal{C}_f
\right)
\right).
\end{aligned}
\label{eq:subject_repr}
\end{equation}
where $z_i$ and $z_t$ are the construct-mean pooled embeddings of the per-construct cues $c_i^f$ in \eqref{eq:cue}, for the source and target subjects, respectively.

We refer to each human-rated source instance admitted to the KB as a source \emph{exemplar}. Each exemplar is stored as an \emph{evidence card} $(z_i,x_i^{\mathrm{display}},y_{i\rightarrow t})$ in the knowledge base. Here $z_i$ is the retrieval representation, $x_i^{\mathrm{display}}$ is the projected textual description shown to the raters (without slot labels), and $y_{i\rightarrow t}$ is the source human-rated profile in the target construct space. Representation Alignment is applied only to the source during KB construction. Target observations are directly converted into the same cue representation.

\subsection{Personality-space Max-Min Evidence Selection}
\label{sec:pmes}

For a target subject, evidence selection balances behavioral relevance with complementary coverage of the human-rated personality space. The target construct space $\mathcal{F}_t$ induces the corresponding source-side evidence space after source-profile mapping. The scores in this space are fixed before target evaluation rather than inferred from target subjects.

Given the target representation $z_t$ in \eqref{eq:subject_repr}, Stage~1 retrieves a candidate pool $C_t$ of $K_{\mathrm{pre}}$ source evidence cards by semantic similarity. Here $\mathrm{sim}(z_t,z_i)$ is the cosine similarity between the target representation and the retrieval representation $z_i$ of card $e_i$. Stage~2 applies the max-min selection within this pool subject to an evidence budget $K \le K_{\mathrm{pre}}$. Candidates with scores for all $f\in\mathcal{F}_t$ form $C_t^{\mathrm{comp}}\subseteq C_t$. For these candidates, the standardized personality-space distance is

\begin{equation}
d_{\sigma}(e_i,e_j)
=
\sqrt{
\sum_{f\in\mathcal{F}_t}
\left(
\frac{s_f^i-s_f^j}{\sigma_f}
\right)^2
},
\label{eq:personality_distance}
\end{equation}
where $s_f^i$ and $s_f^j$ denote the mapped source scores of cards $e_i$ and $e_j$ on construct $f$, and $\sigma_f$ is the standard deviation of construct $f$ computed from complete source-KB profiles.

The first selected card is the semantically nearest complete-profile candidate. Each subsequent card greedily maximizes its minimum personality-space distance to the selected set:

\begin{equation}
e^{(k)} =
\arg\max_{e_i\in C_t^{\mathrm{comp}}\setminus E}
\min_{e_j\in E}
d_{\sigma}(e_i,e_j),
\label{eq:maxmin_selection}
\end{equation}
where $e^{(k)}$ is the $k$-th selected card and $E$ is the set of cards already selected in earlier iterations.

Semantic similarity defines the relevant behavioral region. Personality-space distance then spreads the human-rated profiles across the target construct space within that region. If fewer than $K$ complete-profile candidates are available, remaining positions are filled by candidates in decreasing semantic-similarity order.

\begin{algorithm}[t]
\caption{Personality-space Max-Min Evidence Selection (PMES)}
\label{alg:pmes}
\begin{algorithmic}[1]
\Require Candidate pool $C_t$; target representation $z_t$;
evidence budget $K$
\Ensure Selected evidence set $E$

\State $e^{(1)} \gets
\displaystyle\arg\max_{e_i\in C_t^{\mathrm{comp}}}
\mathrm{sim}(z_t,z_i)$
\State $E \gets \{e^{(1)}\}$

\While{$|E|<K$ and
$C_t^{\mathrm{comp}}\setminus E\neq\emptyset$}
\State $e^{(|E|+1)} \gets
\displaystyle\arg\max_{e_i\in C_t^{\mathrm{comp}}\setminus E}
\left[
\displaystyle\min_{e_j\in E}
d_{\sigma}(e_i,e_j)
\right]$
\State $E\gets E\cup\{e^{(|E|+1)}\}$
\EndWhile

\State Append remaining candidates in decreasing semantic similarity
order if $|E|<K$

\State \Return $E$
\end{algorithmic}
\end{algorithm}

The standard deviations in \eqref{eq:personality_distance} are
computed solely from complete source-KB profiles. The greedy selection in \eqref{eq:maxmin_selection} also operates only on complete-profile candidates. PMES does not need target personality labels or target-side personality statistics.

\subsection{Evidence-Grounded Continuous Scoring}
\label{sec:scoring}

We reuse the construct space $\mathcal{F}_t$ defined in the problem setting. The multi-rater scorer, hereafter \textbf{Cont3}, produces a continuous score vector $\hat{\mathbf{y}}_t \in [1,5]^{|\mathcal{F}_t|}$ over the constructs in $\mathcal{F}_t$. When the target measures are coarser, a predefined aggregation of the corresponding construct scores is applied consistently to the predictions and the ground-truth targets.

Each independent LLM rater receives the target construct definitions and assessment scenarios, the selected evidence cards with their behavioral descriptions and human-rated profiles, and the target subject's multimodal behavioral description. For construct $f$, rater $m$ produces a continuous score $y_f^{(m)} \in [1,5]$.

We instantiate Cont3 with three independent raters, each based on a different LLM. A rater is valid only if it returns a complete score vector over all constructs in $\mathcal{F}_t$. At least two valid raters are required for a subject to be retained. Subjects with fewer valid raters are excluded from evaluation. The final construct-level prediction averages the valid judgments:

\begin{equation}
\hat{y}_f
=
\frac{1}{|\mathcal{M}|}
\sum_{m\in\mathcal{M}}
y_f^{(m)},
\label{eq:score_mean}
\end{equation}
where $\mathcal{M}$ is the set of valid raters. Raters output only the requested continuous scores. No intermediate chain-of-thought or other rater reasoning is used in aggregation.

\section{Experimental Setup}
\label{sec:setup}

\subsection{Datasets}

We evaluate the framework on two separately collected asynchronous video interview (AVI) datasets, \textbf{OPVA}~\cite{zhang2026} and \textbf{AVI-6}~\cite{avi6}. The two datasets differ in question design and in annotation granularity, the two axes that jointly define the cross-protocol transfer setting. AVI-6 provides factor-level annotations over Honesty-Humility (H), Extraversion (X), Agreeableness (A), and Conscientiousness (C). OPVA carries facet-level annotations that are aggregated to the factor level through the predefined target annotation mapping. Both datasets retain the construct annotations needed to map a source profile into the target construct space. The transfer directions differ only in their evaluation space (Table~\ref{tab:transfer_setup}). The two datasets therefore allow evaluation of the same evidence construction, retrieval, and scoring framework in both transfer directions. For each direction, source-side personality profiles are mapped into the target construct space before knowledge-base indexing. Sample-level filtering is applied as described in the Supplementary Material.

\begin{table}[t]
\centering
\caption{Bidirectional cross-protocol transfer settings.
\emph{Facet-mean} denotes facet-level targets aggregated to the
factor level.}
\label{tab:transfer_setup}
\begin{tabularx}{\columnwidth}{l c c X c}
\toprule
Direction & Source & Target & Evaluation space & $n$ \\
\midrule
Forward & OPVA & AVI-6 & H/X/A/C & 644 \\
Reverse & AVI-6 & OPVA & X/C (facet-mean) & 677 \\
\bottomrule
\end{tabularx}
\end{table}

\subsection{Implementation Details}
\label{sec:implementation}

All experiments use the method described in Section~\ref{sec:method}. Retrieval configurations are fixed before target evaluation. 

\paragraph{Preprocessing}
Visual descriptions are generated with Qwen2.5-VL-7B-Instruct~\cite{bai2025}, and audio descriptions with Fun-Audio-Chat-8B~\cite{team2026}. Dataset transcripts are used directly. Perception prompts restrict outputs to observable behavior. Decoding, sampling, and prompt specifications are provided in the Supplementary Material.

\paragraph{Representation Alignment}
Cue texts are generated with \texttt{Qwen3-Max-2026-01-23} and encoded with Octen-Embedding-0.6B~\cite{octen2025rteb}. Subject representations average the corresponding construct-level cue vectors. Protocol Projection uses \texttt{Qwen3.5-Plus}, distinct from the cue-generation and scoring models. The closed-source models are used only for representation construction. Personality scoring uses the open-source raters described below. \texttt{Qwen3.5-Plus} also appears as a zero-shot textual baseline in Table~\ref{tab:external_main}. Temperature, decoding, and embedding-dimension settings are provided in the Supplementary Material.

\paragraph{Retrieval}
Retrieval is fixed before target evaluation at $(K_{\mathrm{pre}},K)=(50,5)$. The configuration is selected based on the evidence budget, personality-space diversity, and context length (see Section~\ref{sec:discussion} for the retrieval-budget analysis). The same configuration is used in both transfer directions without target-side retuning.

\paragraph{Multi-rater scoring}
We use three open-source LLMs as independent continuous raters: Qwen3-14B~\cite{qwen3}, Llama-3.1-8B-Instruct~\cite{grattafiori2024llama3herdmodels}, and Gemma-3-12B-it~\cite{gemmateam2025gemma3technicalreport}. All are served with vLLM at temperature 0 and produce scores in $[1.0,5.0]$ for all constructs required by the target assessment protocol. Subjects with fewer than two complete rater outputs are excluded, and valid ratings are averaged as defined in Section~\ref{sec:scoring}. The same three models are used individually in the Textualized LLM zero-shot baselines (Table~\ref{tab:external_main}). The evidence-free ablation applies the Cont3 ensemble without retrieved evidence.

\subsection{Evaluation Metrics}
\label{sec:metrics}

We report mean squared error (MSE) in the original $[1,5]$ score space (lower is better). MSE is the primary metric because the task is continuous score prediction and MSE measures squared prediction error directly. Mean absolute error (MAE) is reported as a complementary magnitude-oriented measure. Its definition and complete tables are provided in the Supplementary Material. Let $N$ denote the number of target subjects, $\mathcal{E}$ the primary evaluation space of the transfer setting, and $y_{if}$ and $\hat{y}_{if}$ the ground-truth and predicted scores of subject $i$ on evaluation construct $f$, respectively. For construct $f$,

\begin{equation}
\mathrm{MSE}_f =
\frac{1}{N}\sum_{i=1}^{N}
(\hat{y}_{if}-y_{if})^2.
\label{eq:mse_f}
\end{equation}

The primary metric reported in the tables is the average across the evaluation constructs:

\begin{equation}
\overline{\mathrm{MSE}} =
\frac{1}{|\mathcal{E}|}\sum_{f\in\mathcal{E}}\mathrm{MSE}_f,
\label{eq:macro_mse}
\end{equation}
where MSE is computed per construct and then averaged across constructs, so that every construct contributes equally. The primary evaluation space is $\mathcal{E}=\{H,X,A,C\}$ for Forward transfer and the aggregated $\mathcal{E}=\{X,C\}$ targets for Reverse transfer. Here $\mathcal{E}$ is the aggregated evaluation view of the target construct space $\mathcal{F}_t$ defined in Section~\ref{sec:method}.

To quantify uncertainty in paired comparisons, we use a paired subject-level bootstrap. For each target subject $i$, we retain the paired predictions produced by Ours and the comparison condition and compute their subject-level squared-error difference
\begin{equation}
d_i =
\frac{1}{|\mathcal{E}|}
\sum_{f\in\mathcal{E}}
\left[
\left(\hat{y}^{\mathrm{Ours}}_{if}-y_{if}\right)^2
-
\left(\hat{y}^{\mathrm{Comp}}_{if}-y_{if}\right)^2
\right].
\label{eq:delta}
\end{equation}
The reported $\Delta$MSE is the mean of $d_i$ over all target subjects. We resample target subjects with replacement and keep the pairing between the two conditions. We then recompute $\Delta$MSE for each bootstrap sample and repeat $B=10{,}000$ times. The 95\% confidence interval is obtained from the 2.5th and 97.5th percentiles of the bootstrap distribution. This interval captures uncertainty in the paired performance difference rather than assuming the two MSE estimates are independent. A negative $\Delta$MSE favors Ours. An interval entirely below zero means the improvement persists across resamples.
\section{Results}
\label{sec:results}

\subsection{Comparison with Baselines}

We first evaluate whether human-rated behavioral evidence can be transferred across independently collected assessment protocols without target-dataset personality supervision. Throughout the section, we call this supervision-free setting \emph{target-label-free} and refer to a method that additionally receives no retrieved evidence as \emph{zero-shot} (ZS).

The baseline set covers two complementary zero-shot settings. The \emph{Textualized LLM} baselines read the same behavioral descriptions as our framework but score personality directly, without retrieved evidence. Because the input text is held fixed, they isolate the contribution of the retrieved-evidence block. This group includes the three open-source raters used by our scorer, plus their evidence-free three-rater ensemble (Ours w/o evidence, the Cont3 configuration). It also includes two closed-source text models (Qwen3.5-Plus and GPT-5). The \emph{Native AV} baselines operate directly on the raw audio or video stream with the perception models used for description generation in preprocessing (Qwen2.5-VL and Fun-Audio-Chat). They test whether skipping textualization and retrieval is sufficient. Our method is evaluated in both directions using the fixed configuration described in Section~\ref{sec:setup}. Table~\ref{tab:external_main} summarizes the primary comparison.

We intentionally exclude target-supervised baselines, such as supervised fine-tuning or target-domain label training. These methods require target personality labels and retraining for each new protocol, so they operate at a higher supervision level than the target-label-free setting studied here. The strongest evidence-free configuration (Ours w/o evidence) is therefore the correct reference for isolating the contribution of retrieved human-rated evidence.

In Forward transfer, Ours achieves a primary MSE of $0.2279$, a $43.8\%$ error reduction relative to the strongest textualized zero-shot baseline (Ours w/o evidence, $0.4052$). In Reverse transfer, it achieves a primary MSE of $0.1651$, a $17.0\%$ reduction relative to the strongest evidence-free baseline (Ours w/o evidence, $0.1990$). This is also the best MSE among target-label-free methods. The complementary MAE results follow the same ranking and are reported in the Supplementary Material.

At the construct level, the improvement is selective rather than uniform. In Forward transfer, Ours attains the lowest MSE on Honesty-Humility, Agreeableness, and Conscientiousness. Qwen3.5-Plus is slightly better on Extraversion (MSE $0.1937$ vs.\ $0.2249$), which is the only Forward construct on which a text-only zero-shot baseline leads the full framework. In Reverse transfer, Ours is lowest on both OPVA constructs under MSE.

\begin{table*}[t]
\centering
\caption{External comparison of the Forward and Reverse transfer
settings (MSE, lower is better). All baselines are zero-shot (ZS) and
receive the same behavioral descriptions without retrieved evidence.
Bold denotes the best (lowest) result within each column.}
\label{tab:external_main}
\resizebox{\linewidth}{!}{%
\begin{tabular}{llcccccccc}
\toprule
\multicolumn{10}{c}{\textbf{Mean Squared Error (MSE)}} \\
\midrule
\multicolumn{2}{c}{Dataset}
& \multicolumn{5}{c}{AVI-6}
& \multicolumn{3}{c}{OPVA} \\
\midrule
Category & Approach & H & X & A & C & Avg & X & C & Avg \\
\midrule
\multirow{6}{*}{Textualized LLM ZS}
& Qwen3-14B & 0.5282 & 0.3919 & 0.5603 & 0.5817 & 0.5155 & 0.2267 & 0.1788 & 0.2027 \\
& Llama-3.1-8B & 0.4315 & 0.3652 & 0.5210 & 0.4934 & 0.4528 & 0.3360 & 0.2612 & 0.2986 \\
& Gemma-3-12B & 0.5273 & 0.3156 & 0.5773 & 0.6215 & 0.5104 & 0.2170 & 0.1984 & 0.2077 \\
& Ours w/o evidence & 0.4380 & 0.2971 & 0.3991 & 0.4867 & 0.4052 & 0.2217 & 0.1764 & 0.1990 \\
& Qwen3.5-Plus & 0.6212 & \textbf{0.1937} & 0.3679 & 0.6371 & 0.4550 & 0.2035 & 0.2396 & 0.2216 \\
& GPT-5~\cite{singh2026openaigpt5card} & 0.7759 & 0.3014 & 0.6961 & 0.8095 & 0.6457 & 0.2243 & 0.2786 & 0.2515 \\
\midrule
\multirow{2}{*}{Native AV ZS}
& Qwen2.5-VL video & 1.0503 & 0.2567 & 0.6467 & 0.5655 & 0.6298 & 1.3123 & 0.5871 & 0.9497 \\
& Fun-Audio-Chat & 2.0098 & 0.4191 & 1.5715 & 1.5931 & 1.3984 & 0.5265 & 0.3671 & 0.4468 \\
\midrule
Ours & Ours (full)
& \textbf{0.1998} & 0.2249 & \textbf{0.2032} & \textbf{0.2837} & \textbf{0.2279} & \textbf{0.1726} & \textbf{0.1576} & \textbf{0.1651} \\
\bottomrule
\end{tabular}}
\end{table*}

Because Ours w/o evidence is the evidence-free version of the scorer used by the full method, we also verify significance against Qwen3.5-Plus, the strongest zero-shot textual baseline outside the Cont3 rater set. All four paired bootstrap intervals lie entirely below zero, so Ours outperforms both the evidence-free ensemble and this external baseline with consistent subject-level significance in both transfer directions (Table~\ref{tab:primary_bootstrap}). Relative to Qwen3.5-Plus, the Forward improvement corresponds to a $49.9\%$
relative error reduction.

\begin{table}[t]
\centering
\caption{Paired subject-level bootstrap comparison against the
evidence-free scorer (Ours w/o evidence) and the strongest zero-shot
textual baseline outside the Cont3 rater set (Qwen3.5-Plus).}
\label{tab:primary_bootstrap}
\scriptsize
\setlength{\tabcolsep}{2.5pt}
\begin{tabular}{@{}llccc@{}}
\toprule
Dir. & Baseline & MSE & $\Delta$MSE & 95\% CI \\
\midrule
\multirow{2}{*}{Fwd.}
& Ours w/o evidence & 0.4052 & $-0.1773$ & $[-0.1990,\,-0.1563]$ \\
& Qwen3.5-Plus
& 0.4550 & $-0.2271$ & $[-0.2508,\,-0.2036]$ \\
\multirow{2}{*}{Rev.}
& Ours w/o evidence & 0.1990 & $-0.0339$ & $[-0.0481,\,-0.0201]$ \\
& Qwen3.5-Plus
& 0.2216 & $-0.0565$ & $[-0.0734,\,-0.0386]$ \\
\bottomrule
\end{tabular}
\end{table}

\subsection{Component Ablation}

We next isolate the contributions of retrieved human-rated evidence, personality-space selection, and Representation Alignment using matched configurations. All comparisons use the same target subjects, target annotations, scoring models, and assessment configuration. Only the component under investigation is changed.

We consider three controlled ablations. \textbf{Ours w/o evidence} removes the retrieved-evidence block while keeping the construct definitions, assessment scenarios, target behavioral descriptions, and the Cont3 multi-rater scorer unchanged. \textbf{Ours w/o PMES} replaces personality-space Max-Min selection with similarity-only Top-$5$ selection from the same Stage-1 candidate pool, which isolates the effect of personality-space complementarity. \textbf{Ours w/o Alignment} removes the entire Representation Alignment pipeline, including Protocol Projection and Cue Construction, from retrieval. Source evidence and target queries then use the corresponding raw behavioral text, while PMES and multi-rater scoring remain unchanged. Because Cue Construction operates on the protocol-projected source representation, Protocol Projection and Cue Construction are ablated jointly rather than evaluated as separate components. All paired comparisons reuse the bootstrap procedure defined in Section~\ref{sec:metrics}.

Ours w/o evidence increases MSE from $0.2279$ to $0.4052$ in Forward transfer and from $0.1651$ to $0.1990$ in Reverse transfer. The paired bootstrap confirms a consistent, subject-level improvement from incorporating retrieved human-rated evidence in both transfer directions (Table~\ref{tab:primary_bootstrap}, Ours w/o evidence rows).

Ours w/o PMES increases MSE to $0.2838$ in Forward transfer and $0.1750$ in Reverse transfer. The paired MSE differences are $-0.0559$ (95\% CI $[-0.0695,-0.0424]$) and $-0.0099$ (95\% CI $[-0.0197,-0.0001]$), respectively. PMES therefore provides a clearer additional gain in Forward transfer and a marginal gain in Reverse transfer, where the CI upper bound is near zero. 

Ours w/o Alignment degrades performance substantially. MSE rises to $0.3636$ in Forward transfer and $0.4235$ in Reverse transfer. The paired differences are $-0.1357$ (95\% CI $[-0.1553,-0.1159]$) and $-0.2584$ (95\% CI $[-0.2835,-0.2333]$). In Reverse transfer this is the largest single-component degradation, whereas in Forward transfer removing retrieved evidence is the largest single-component loss ($0.4052$).

Taken together, these results show distinct contributions from the three components. Retrieved human-rated evidence provides the primary gain over evidence-free scoring in Forward transfer. Representation Alignment provides the primary gain in Reverse transfer and is necessary for cross-protocol comparability in both directions. PMES provides a complementary gain by making the retrieved evidence set more diverse.

\begin{table*}[t]
\centering
\caption{Component ablation results across Forward and Reverse transfer
(MSE, lower is better). All retrieval-based configurations use Cont3
with $(K_{\mathrm{pre}},K)=(50,5)$. Bold denotes the best (lowest)
result within each column.}
\label{tab:main_ablation}
\resizebox{\linewidth}{!}{%
\begin{tabular}{llcccccccc}
\toprule
\multicolumn{10}{c}{\textbf{Mean Squared Error (MSE)}} \\
\midrule
\multicolumn{2}{c}{Dataset}
& \multicolumn{5}{c}{AVI-6}
& \multicolumn{3}{c}{OPVA} \\
\midrule
Category & Configuration & H & X & A & C & Avg & X & C & Avg \\
\midrule
\multirow{2}{*}{Evidence retrieval}
& Ours w/o evidence & 0.4380 & 0.2971 & 0.3991 & 0.4867 & 0.4052 & 0.2217 & 0.1764 & 0.1990 \\
& Ours w/o PMES & 0.2084 & 0.2728 & \textbf{0.1952} & 0.4586 & 0.2838 & 0.1766 & 0.1734 & 0.1750 \\
\midrule
Representation Alignment
& Ours w/o Alignment & 0.2517 & 0.3980 & 0.2847 & 0.5199 & 0.3636 & 0.3465 & 0.5005 & 0.4235 \\
\midrule
Full model
& Ours (full) & \textbf{0.1998} & \textbf{0.2249} & 0.2032 & \textbf{0.2837} & \textbf{0.2279} & \textbf{0.1726} & \textbf{0.1576} & \textbf{0.1651} \\
\bottomrule
\end{tabular}}
\end{table*}

\section{Discussion}
\label{sec:discussion}

The results raise two questions about cross-protocol evidence transfer. The first concerns what actually transfers across assessment protocols. A retrieved source record carries observed behavior plus the human personality judgment attached to that behavior. The two need not transfer in the same way. If the personality label transfers rather than the behavior-to-personality association, cross-protocol prediction reduces to nearest-neighbor label copying, where a behaviorally similar source subject is treated as though its personality profile were equally close to that of the target. If only the empirical association transfers, the retrieved record is evidence to interpret rather than a surrogate label. The second concerns how robustly the transferred evidence can be interpreted once it is identified. We examine the two questions with two sets of complementary analyses. The first set tests label transfer directly and diagnoses how the evidence set should be built. The second set decomposes the evidence-grounded scorer into individual models and measures inter-rater reliability.

\subsection{What Transfers, and How Should the Evidence Set Be Built?}

Whether a retrieved record should be copied as a label or interpreted as evidence determines how the assessment is designed. The label account is the simpler one, so we rule it out first by replacing evidence-grounded scoring with the simplest possible label transfer. We average the human-rated profiles attached to the retrieved records, which are the Protocol-Projected profiles mapped into the target construct space without any LLM reasoning or statistical rescaling. We compare two neighbor sets. The similarity-only Top-5 set holds the behaviorally most similar records, the set a nearest-neighbor label copier would use. The PMES set is the complementary evidence set the method actually uses. Table~\ref{tab:label_transfer} compares Cont3 with averaging the same records' personality labels. All four subject-level $\Delta$MSE intervals exclude zero, so on either retrieval set treating the records as evidence outperforms treating their scores as transferable labels. Diversifying the evidence has a direction-dependent effect on label averaging, improving the Forward estimate and worsening it in Reverse. Diversification does not reliably rescue label transfer. Behaviorally similar source subjects are therefore not reliable surrogate labels for the target. A personality rating is attached to an individual observed under a particular protocol. Behavioral similarity across protocols does not guarantee an equivalent underlying profile. The retrieved record is useful as empirical evidence about how human raters have interpreted a behavioral pattern. The final prediction remains a judgment about the target.

\begin{table}[t]
\centering
\caption{Evidence-grounded Cont3 vs.\ neighbor-label mean under matched
retrieval sets ($K{=}5$). Negative $\Delta$MSE favors Cont3.}
\label{tab:label_transfer}
\scriptsize
\setlength{\tabcolsep}{2pt}
\begin{tabular}{@{}llcc@{}}
\toprule
Retrieval & Predictor & Fwd. & Rev. \\
\midrule
\multirow{3}{*}{Sim-$5$}
& Cont3 & 0.2838 & 0.1750 \\
& Neighbor & 0.3550 & 0.3341 \\
& $\Delta$MSE & $-0.0713$ & $-0.1591$ \\
\multicolumn{4}{@{}l@{}}{\scriptsize CI:
$[-.0839,-.0588]$;\ $[-.1783,-.1401]$} \\
\midrule
\multirow{3}{*}{PMES}
& Cont3 (Ours) & \textbf{0.2279} & \textbf{0.1651} \\
& Neighbor & 0.2905 & 0.3792 \\
& $\Delta$MSE & $-0.0626$ & $-0.2141$ \\
\multicolumn{4}{@{}l@{}}{\scriptsize CI:
$[-.0774,-.0481]$;\ $[-.2363,-.1923]$} \\
\bottomrule
\end{tabular}
\end{table}
This interpretation also clarifies the role of PMES. If retrieved records are evidence rather than labels, retrieval should build a set whose members provide complementary information instead of finding the most behaviorally similar subjects. Table~\ref{tab:retrieval_diagnostics} shows the trade-off. The reported minimum $d_{\sigma}$ is the minimum pairwise standardized personality-space distance within the selected evidence set, averaged over queries. PMES maximizes this quantity, which therefore serves as the primary diversity statistic. Enlarging the candidate pool raises the minimum pairwise $d_{\sigma}$ within the selected records, while enlarging the evidence budget lowers it, because more records are packed into a bounded selection rather than adding redundant information. We therefore fix $(K_{\mathrm{pre}},K)=(50,5)$ on engineering grounds. This choice is not tuned to downstream prediction error, as doing so would compromise the zero-shot setting. It is also not a claim that a larger evidence budget is always better. Because the budget is a single global setting applied identically in both directions, we report the diagnostic on the Forward queries. The trade-off concerns the construction of the evidence set rather than its interpretation.

Cross-protocol retrieval should therefore be formulated as evidence construction rather than nearest-neighbor label transfer. The retrieved records matter because they jointly carry complementary information about how observed behavior is interpreted, not because any one of them is behaviorally similar to the target.

\begin{table}[t]
\centering
\caption{Forward offline retrieval-budget diagnostics ($n{=}644$).}
\label{tab:retrieval_diagnostics}
\begin{tabular}{lccc}
\toprule
\multicolumn{4}{c}{(a) Candidate-pool size (fixed $K{=}5$)} \\
\midrule
$K_{\mathrm{pre}}$ & 20 & 50 & 100 \\
Minimum $d_{\sigma}$ & 2.08 & 2.54 & 2.85 \\
\midrule
\multicolumn{4}{c}{(b) Evidence budget (fixed $K_{\mathrm{pre}}{=}50$)} \\
\midrule
$K$ & 3 & 5 & 10 \\
Minimum $d_{\sigma}$ & 3.19 & 2.54 & 1.82 \\
Context (tokens) & $\sim$8.4k & $\sim$9.6k & $\sim$12.6k \\
\bottomrule
\end{tabular}
\end{table}

\subsection{How Robustly Can Transferred Evidence Be Interpreted?}

Even with a useful evidence set, the final assessment depends on the model that interprets it. Model-specific variation is a distinct failure mode from evidence-selection error. We therefore separate two quantities that are easy to conflate: prediction error (MSE) and inter-rater reliability (ICC). Reliability asks whether the scoring process is stable across independent models, and error asks whether the prediction is close to the target. Aggregation is not a uniformly stronger predictor. In Forward transfer, the mean of the three individual raters, denoted Cont3, is clearly better than the best single model. In Reverse transfer, Cont3 is essentially tied with the strongest individual rater (Qwen3-14B). The direction-dependent benefit of averaging motivates measuring reliability rather than error alone.

Reliability is quantified with the Shrout-Fleiss intraclass correlation under a two-way random-effects model with absolute agreement. It is computed separately for each construct (facet in Reverse) from the three scoring models' per-subject predictions. The single-rater reliability is
\begin{equation}
\mathrm{ICC}(2,1)=\frac{\mathrm{MS}_{\mathrm{B}}-\mathrm{MS}_{\mathrm{E}}}
{\mathrm{MS}_{\mathrm{B}}+(k-1)\,\mathrm{MS}_{\mathrm{E}}
+\frac{k}{n}\,(\mathrm{MS}_{\mathrm{J}}-\mathrm{MS}_{\mathrm{E}})},
\label{eq:icc21}
\end{equation}
where $\mathrm{MS}_{\mathrm{B}}$ is the between-subjects mean square, $\mathrm{MS}_{\mathrm{J}}$ the between-raters mean square, $\mathrm{MS}_{\mathrm{E}}$ the residual mean square, $k=3$ the number of raters, and $n$ the number of subjects. The reliability of the three-rater mean follows from the Spearman--Brown formula
\begin{equation}
\mathrm{ICC}(2,3)=\frac{3\,\mathrm{ICC}(2,1)}{1+2\,\mathrm{ICC}(2,1)}.
\label{eq:icc23}
\end{equation}
In Forward transfer, ICC(2,1) ranges from $0.272$ to $0.678$ and ICC(2,3) from $0.529$ to $0.863$. In Reverse the corresponding ranges are $0.500$ to $0.834$ and $0.750$ to $0.938$. Median ICC(2,1) is higher in Reverse than in Forward, but the construct-wise ranges still overlap. We treat these as descriptive only because Forward and Reverse use different construct granularities. The low end of the Forward range indicates that for some constructs the three models agree only weakly, so agreement is not uniformly high in either direction.

Pairwise Pearson correlations among the three raters range from $0.532$ to $0.743$ in Forward and $0.746$ to $0.803$ in Reverse, separating the two directions more cleanly than the overlapping ICC ranges. Both sets of statistics are point estimates without confidence intervals or a formal between-direction test. In Forward, raters disagree more strongly and averaging has more variation to attenuate, which match the larger MSE gain. In Reverse, raters already agree more and the additional gain is small. Reliability and prediction error are therefore complementary: MSE evaluates closeness to the target rating, while ICC characterizes the stability of the scoring process.

Taken together, the two sets of analyses support cross-protocol personality assessment as empirical evidence transfer rather than direct label transfer, with retrieval framed as evidence construction and multi-rater aggregation as a robustness mechanism.

\section{Limitations}
\label{sec:limitations}

Two limitations bound the current results.

First, building the source knowledge base introduces an information bottleneck. Protocol Projection and Cue Construction convert multimodal behavioral observations into target-protocol-compatible semantic text. This enables cross-protocol retrieval but trades off modality-specific signals and fine-grained temporal structure. Retaining raw video and audio would reduce this loss but sharply increase retrieval cost. Future work should explore adaptive multimodal representations that use compact semantic representations for efficient retrieval while retaining raw or temporally localized evidence.

Second, the current evaluation covers two independently collected interview datasets and one set of scoring models. Bidirectional transfer provides evidence beyond a single direction, but broader evaluation across protocols, participant populations, and model families is needed to establish generality. A supervised upper bound is not reported, so the performance cost of forgoing target supervision remains unquantified.

\section{Conclusion}
\label{sec:conclusion}

This work addressed cross-protocol personality assessment: reusing human-rated behavioral evidence collected under one protocol to score personality under another without target-dataset supervision. We proposed CEPA (Cross-protocol Evidence-grounded Personality Assessment), a retrieval-augmented framework that aligns multimodal records to the target protocol, builds a compact evidence pool, and interprets that evidence through multi-rater scoring.

Across bidirectional transfer between OPVA and AVI-6, CEPA outperforms direct label transfer, confirming that cross-protocol assessment is feasible without target labels.

Two principles emerge. First, cross-protocol transfer requires behavioral comparability rather than direct personality-label transfer. Second, the value of retrieved evidence depends on complementary organization and joint interpretation under a limited budget. What transfers is the observed behavior-to-judgment association, not the source label itself. Extending CEPA to broader protocols, participant populations, and model families is a natural next step.

\bibliographystyle{IEEEtran}
\bibliography{main}
\vskip -2\baselineskip plus -1fil

\begin{IEEEbiography}
[{\includegraphics[width=1in,height=1.25in,clip,keepaspectratio]{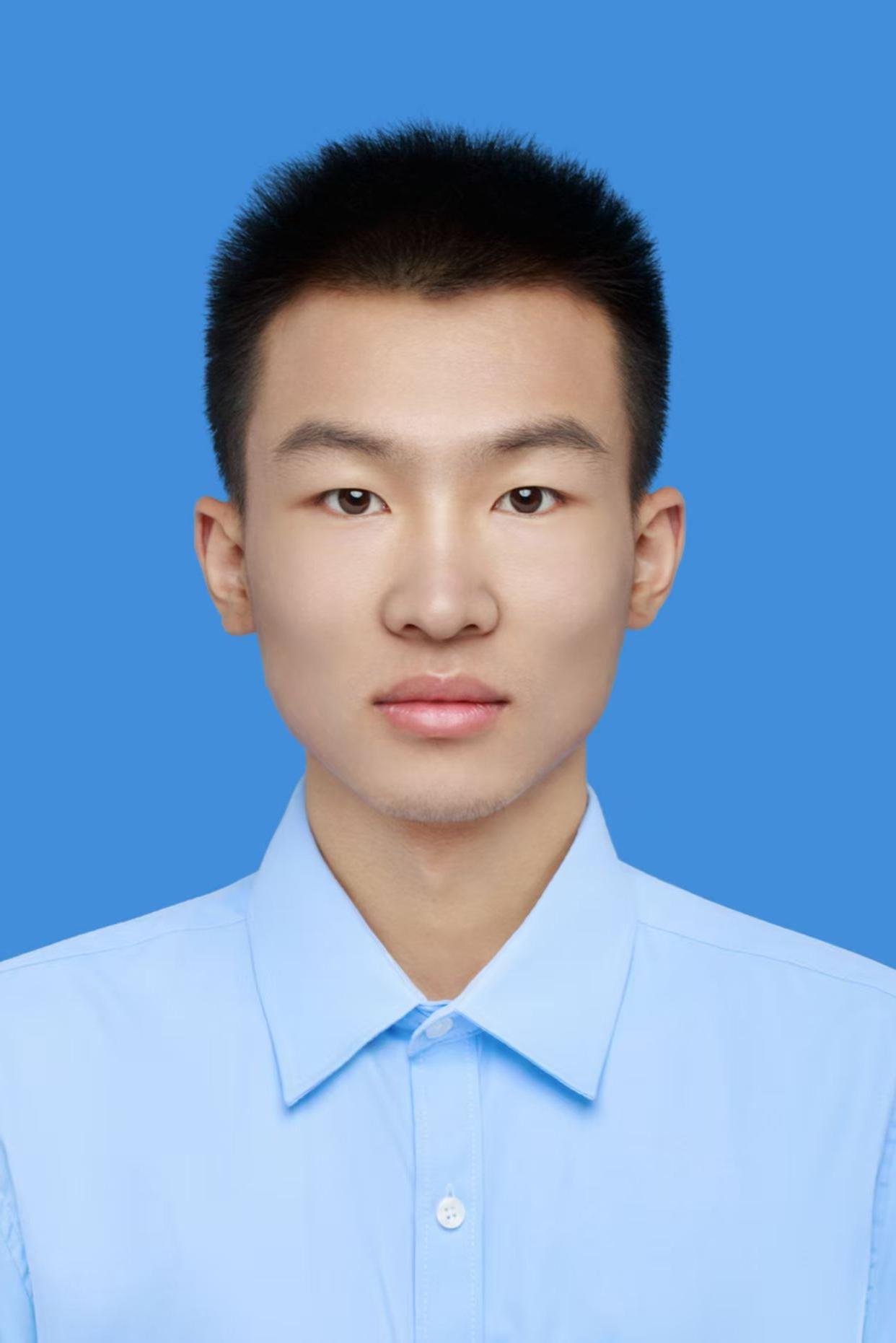}}]{Wei Shan} received the bachelor's degree Nanjing Forestry University in 2024. He is currently pursuing the master's degree with the Southeast University-Monash University Joint Graduate School. His current research interests include large language models , data mining, and personality analysis.
\end{IEEEbiography}
\vskip -2\baselineskip plus -1fil

\begin{IEEEbiography}
[{\includegraphics[width=1in,height=1.25in,clip,keepaspectratio]{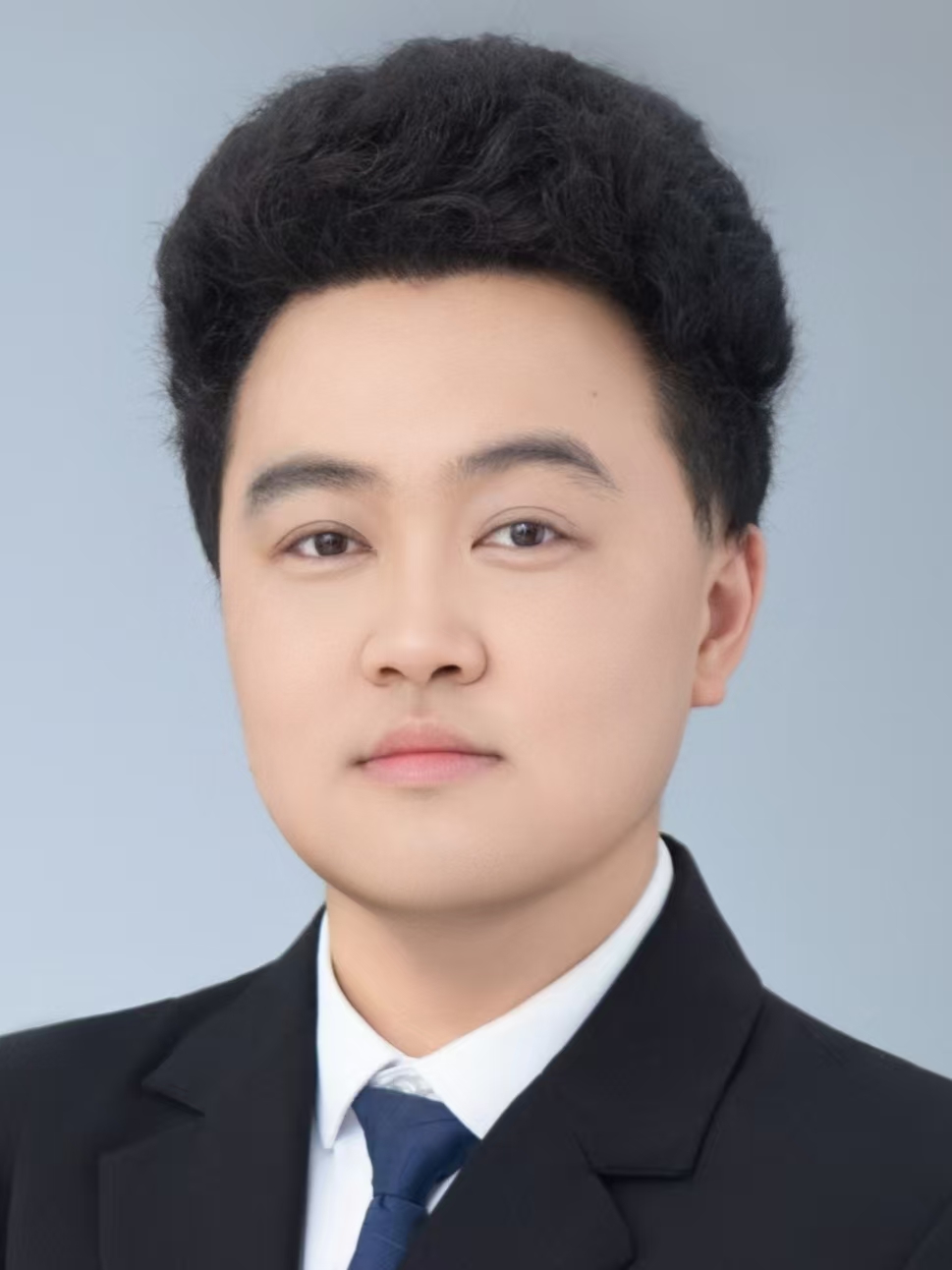}}]{Tianyi Zhang (Senior Member, IEEE)} is an associate professor at the School of Biological Sciences and Medical Engineering, Southeast University. Before that, he worked as a postdoc researcher at Vrije Universiteit Amsterdam. He got his PhD degree in Delft University of Technology. He was also associated with the Distributed \& Interactive Systems (DIS) group at the national research institute for mathematics and computer science in the Netherlands (CWI). His research interests lie in affective computing and personality recognition. He is an Associate Editor of \textit{IEEE Transactions on Affective Computing}. 
\end{IEEEbiography}
\vskip -2\baselineskip plus -1fil
\begin{IEEEbiography}
[{\includegraphics[width=1in,height=1.25in,clip,keepaspectratio]{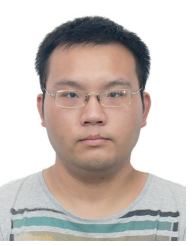}}]
{Yuan Zong (Member, IEEE)} received the Ph.D. degree in biomedical engineering from the Southeast University, Nanjing, China, in 2018. He is currently an Associate Professor with the Key Laboratory of Child Development and Learning Science of Ministry of Education, Southeast University. From 2016 to 2017, he was a Visiting Student with the Center for Machine Vision and Signal Analysis, University of Oulu, Oulu, Finland. His research interests include affective computing, pattern recognition, computer vision, and speech signal processing.
\end{IEEEbiography}
\vskip -2\baselineskip plus -1fil

\begin{IEEEbiography}
[{\includegraphics[width=1in,height=1.25in,clip,keepaspectratio]{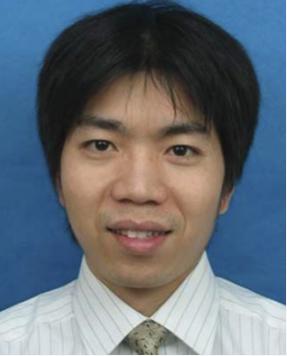}}]
{Wenming Zheng (Senior Member, IEEE)} received the Ph.D. degree in signal processing from Southeast University. Since 2004, he has been with the Research Center for Learning Science, Southeast University. He is currently a Full Professor with the Key Laboratory of Child Development and Learning Science, Ministry of Education, Southeast University. His research interests include affective computing, pattern recognition, machine learning, and computer vision. He has been elected as an IET Fellow since 2022. He is an Associate Editor of \textit{IEEE Transactions on Affective Computing} and the Editorial Board Member of the \textit{Visual Computer}.
\end{IEEEbiography}

\end{document}